\pdfoutput=1

\documentclass[11pt]{article}

\usepackage[final]{acl}

\usepackage{times}
\usepackage{latexsym}

\usepackage[T1]{fontenc}

\usepackage[utf8]{inputenc}

\usepackage{microtype}

\usepackage{inconsolata}

\usepackage{times}
\usepackage{soul}
\usepackage{url}
\usepackage[utf8]{inputenc}
\usepackage{graphicx}
\usepackage{amsmath}
\usepackage{amsthm}
\usepackage{booktabs}
\usepackage{algorithm}
\usepackage[switch]{lineno}

\usepackage{fancyhdr}
\usepackage{balance}
\usepackage{caption}
\usepackage{enumitem}
\usepackage{multicol}
\usepackage{multirow}
\usepackage{subfigure}
\usepackage{textcomp}
\usepackage{threeparttable}
\usepackage{tikz}
\usepackage{url}
\usepackage{xcolor}
\usepackage{xspace}
\usepackage{bbding}
\usepackage{float}
\usepackage{makecell}
\usepackage[]{footmisc}
\usepackage{graphicx}
\usepackage{subcaption} 
\usepackage{xcolor}
\usepackage{algorithm}
\usepackage{algpseudocode}
\usepackage[inkscapelatex=false, notransparent]{svg}
\usepackage{ulem}

\usepackage{grffile}

\DeclareMathOperator*{\argmin}{argmin}

\newcommand{\ignore}[1]{{}}

\newcommand{\eg}{\textit{e.g.}\xspace}

\newcommand{\etal}{\textit{et al.}\xspace}

\begin{document}

\pagestyle{plain}

\title{LLM Factoscope: Uncovering LLMs' Factual Discernment through Measuring Inner States}



\author{
 \textbf{Jinwen He\textsuperscript{1,2}},
 \textbf{Yujia Gong\textsuperscript{1,2}},
 \textbf{Zijin Lin\textsuperscript{1,2}},
 \textbf{Cheng'an Wei\textsuperscript{1,2}},
\\
 \textbf{Yue Zhao\textsuperscript{1,2}},
 \textbf{Kai Chen\textsuperscript{1,2}}
\\
\\
 \textsuperscript{1}Institute of Information Engineering,  Chinese Academy of Sciences,
\\
 \textsuperscript{2}School of Cyber Security, University of Chinese Academy of Sciences,
\\
 \small{
   \textbf{Correspondence:} \href{mailto:zhaoyue@iie.ac.cn}{zhaoyue@iie.ac.cn}   \href{mailto:chenkai@iie.ac.cn}{chenkai@iie.ac.cn}
 }
}

\maketitle
 
\begin{abstract}
Large Language Models (LLMs) have revolutionized various domains with extensive knowledge and creative capabilities. However, a critical issue with LLMs is their tendency to produce outputs that diverge from factual reality. This phenomenon is particularly concerning in sensitive applications such as medical consultation and legal advice, where accuracy is paramount. 
Inspired by human lie detectors using physiological responses, we introduce the LLM Factoscope, a flexible and extendable pipeline that leverages the inner states of LLMs for factual detection. Our investigation reveals distinguishable patterns in LLMs' inner states when generating factual versus non-factual content. We demonstrate its effectiveness across various architectures, achieving over 96\% accuracy on our custom-collected factual detection dataset. Our work opens a new avenue for utilizing LLMs' inner states for factual detection and encourages further exploration into LLMs' inner workings for enhanced reliability and transparency.
\end{abstract}
\section{Introduction}

Large Language Models (LLMs) have gained immense popularity, revolutionizing various domains with their remarkable creative capabilities and vast knowledge repositories. These models reshape fields like natural language processing~\cite{nlp}, content generation~\cite{content}, and more. However, despite their advanced abilities, a growing concern surrounds their propensity for ``hallucination'' — the generation of outputs that deviate from factual reality~\cite{hallucination}. 
In critical applications like medical consultation~\cite{huatuogpt}, legal advice~\cite{chatlaw}, and educational tutoring~\cite{tutoring}, factual LLM outputs are not just desirable but essential, as non-factual outputs from these models could potentially lead to negative outcomes for users, affecting their health, legal standing, or educational understanding. Recognizing this, LLM-generated content's factual detection has emerged as an area of paramount importance~\cite{distinguishinpaper}.
Current research predominantly relies on cross-referencing LLM outputs with external databases~\cite{hallucination}. While effective, this approach necessitates extensive external knowledge bases and sophisticated cross-referencing algorithms, introducing more complexity and dependency. This raises a compelling question: Could we possibly exclude external resources but only leverage the inner states of LLMs for factual detection?

Inspired by human lie detectors, we explore establishing a lie detector for LLMs. Humans show specific physiological changes when making statements that contradict their knowledge or beliefs, enabling lie detection through these changes.
We assume that exposure to a wide range of world knowledge during LLMs' training establishes their knowledge or beliefs.
Moreover, training datasets usually have more factual than non-factual sources~\cite{llama}.
This assumption makes establishing a lie detector for LLMs possible. To achieve this, we need to address two essential questions: (1) What might the ``physiological'' indicators for LLMs be, similar to heart rate and eye movements in humans? (2) How can we leverage these indicators in LLMs to establish the lie detector? However, the lack of interpretability in LLMs makes it impossible for humans to understand model parameters or hidden states, rendering the solutions to these questions challenging.

We address the above two challenges through heuristic and data-driven methods, respectively. For the first challenge, we initially use \textit{activation maps} as static features to represent the ``physiological'' states of the LLM. 
Based on our observations, factual-related activations show higher intensity when the LLM produces factual outputs, making the activation maps representative to serve as features for the lie detector.
Subsequently, we use \textit{final output ranks}, \textit{top-k output indices}, and \textit{top-k output probabilities} across layers as dynamic features to represent the LLM's decision-making processes. 
We note that factual outputs by LLMs show more stable final output ranks, greater semantic similarity in top-k output indices, and larger top-k output probability differences in later layers (See Section~\ref{sec:observation}).
For the second challenge, due to the intricate relationship between the features and the factuality of outputs, we employ a data-driven approach to extract underlying principles from these features. Consequently, we design a systematic and automated factual detection pipeline, LLM Factoscope, that includes the factual datasets collection, the inner states collection, and the design of the factual detection model (See Section~\ref{sec:method}).

In our experiments, we empirically demonstrate the effectiveness of the LLM Factoscope across various LLM architectures, including GPT2-XL-1.5B, Llama2-7B, Vicuna-7B, Stablelm-7B, Llama2-13B, and Vicuna-13B. The LLM Factoscope achieves an accuracy rate exceeding 96\% in factual detection. Additionally, we extensively examine the model's generalization capabilities and conduct ablation studies to understand the impact of different components and parameters on the LLM Factoscope's performance. 
Our work paves a new path for utilizing inner states from LLMs for factual detection, sparking further exploration and analysis of LLMs' inner states for enhanced model understanding and reliability.

Our contributions are as follows: (1) We identify effective inner states closely associated with the factual accuracy of the content generated by LLMs, discovered through observation. (2) We introduce the LLM Factoscope, a novel and extendable pipeline for automating the detection of factual accuracy in LLM outputs, facilitating the incorporation of new datasets with ease. (3) Our model has been empirically validated across various LLM architectures, demonstrating over 96\% accuracy on our custom-collected factual detection dataset. Datasets and code are released for further research: https://github.com/JenniferHo97/llm\_factoscope.

\section{Related Work}


\begin{figure*}[htbp]
\setlength{\abovecaptionskip}{4pt}
\setlength{\belowcaptionskip}{0pt}
\begin{minipage}{0.64\linewidth}
\centering
\includegraphics[width=0.95\linewidth, trim=7 680 0 0,clip]{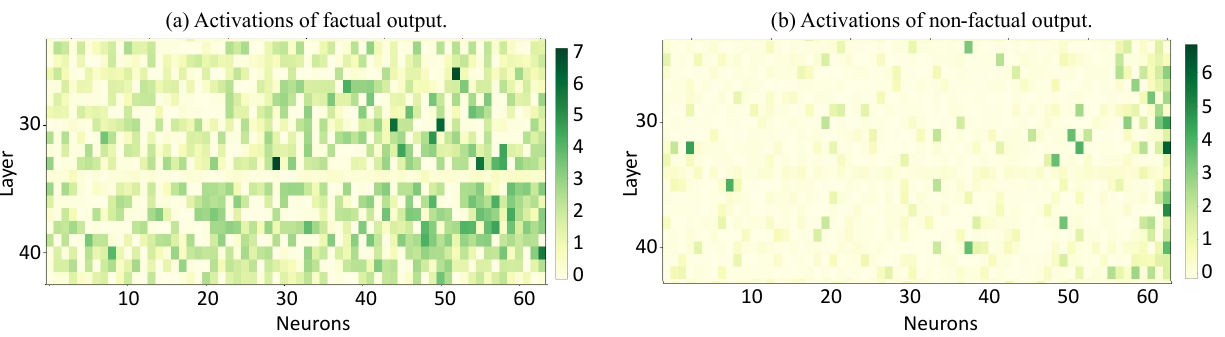}
\caption{Activation maps of layers 24-42, showing the top 1\% factual-related neurons.}
\label{fig:case_activation}
\end{minipage}
\hspace{.15in}
\begin{minipage}{0.32\linewidth}
\centering
\includegraphics[width=0.99\linewidth, trim=10 0 0 18,clip]{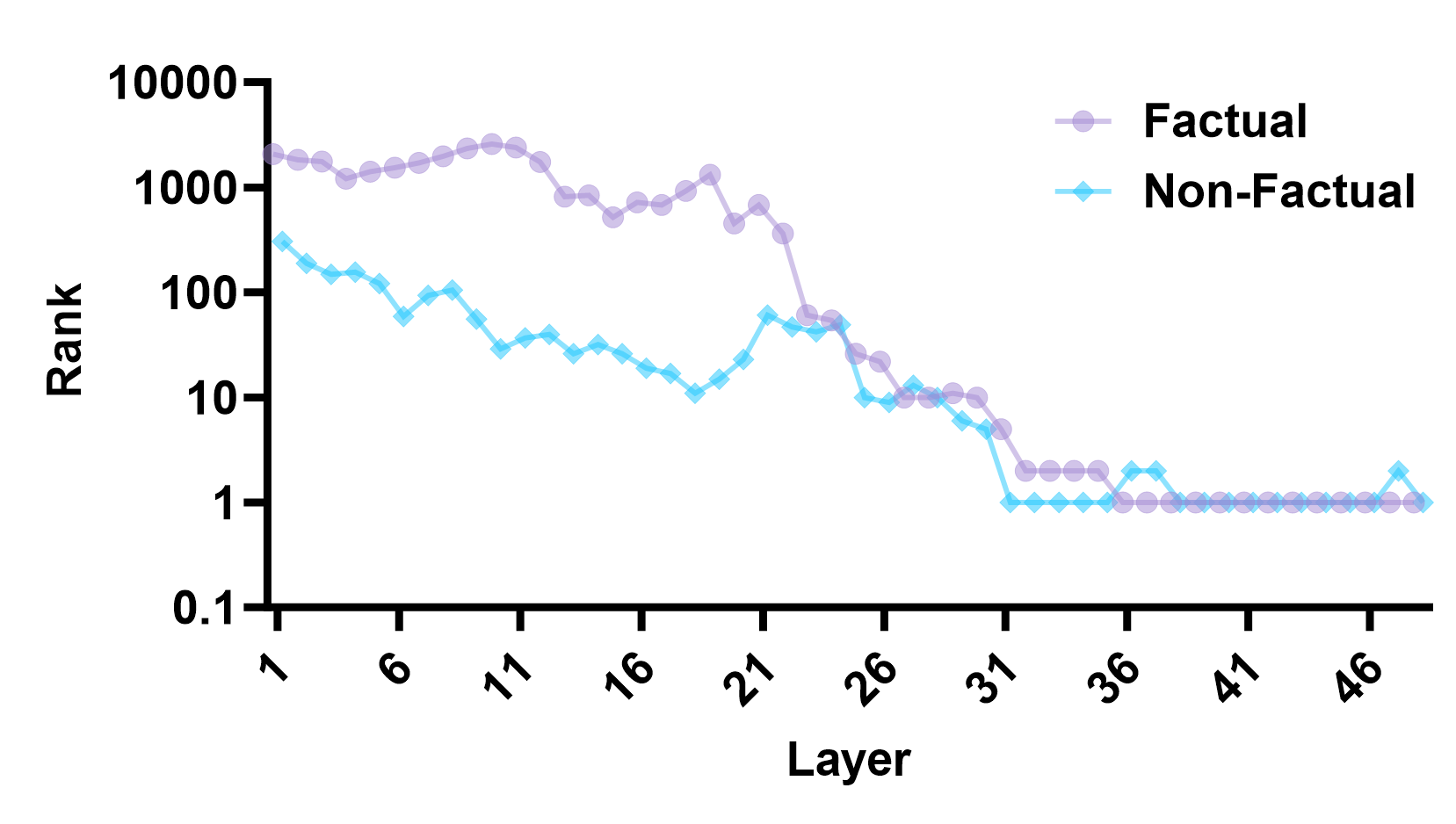}
\caption{Case study of final output rank.}
\label{fig:case_rank}
\end{minipage}
\vspace{-4mm}
\end{figure*}

\subsection{The Logit Lens}

The logit lens, which uses the model’s own unembedding matrix to decode intermediate layers~\cite{logit_lens}, offers valuable insights into the model's decision-making process and enables corrections of harmful behaviors. Geva~\etal\cite{geva} analyzed the decoding output of intermediate layers and extracted human-understandable concepts learned by the parameter vectors of the transformer's feed-forward layers, allowing for manipulation of model outputs to reduce harmful content. Similarly, McGrath ~\etal~\cite{mcgrath} found that late layer MLPs often act to reduce the probability of the maximum-likelihood token, and this reduction can be self-corrected by knocking out certain attention layers. Halawi~\etal~\cite{halawi} decoded intermediate layers to understand the mimicry of harmful outputs by language models, identifying significant shifts in decision-making between correct and incorrect input statements in the mid to later layers, and proposed nullifying key layers to improve output accuracy when facing misleading inputs. These studies reveal the potential of analyzing inner states for improving LLM outputs. In our study, we use the logit lens to decode the outputs of intermediate layers, capturing changes in output indices and probabilities within the LLM.

\subsection{LLM Factual Detection}

Fact-checking LLM outputs is increasingly critical. Current approaches, such as manual examination of training datasets~\cite{examinate_training} and referencing external databases~\cite{Retrieval_Augmentation}, are labor-intensive and computationally expensive. Fact-checking involves detecting the factuality of both inputs and outputs of LLMs. Our task focuses on output factuality detection. 
Previous methods like SPALMA~\cite{internal} and Marks~\etal~\cite{geometry} primarily detect input factuality, while ITI~\cite{ITI} enhances output factuality through inner state manipulation. CCS~\cite{ccs} evaluates LLM outputs using activation values and auto-labeling by LLMs, which we found unreliable. Semantic uncertainty~\cite{semanticuncertainty} analyzes output uncertainty through computationally intensive repeated sampling. Hallucination detection methods like SelfCheckGPT~\cite{selfcheckgpt} and INSIDE~\cite{inside} also rely on sampling, introducing significant computational overheads. 
Our LLM Factoscope enables real-time inference by leveraging various inner states and using a larger, more diverse dataset to examine changes across layers, significantly boosting practicality, accuracy, and generalizability.

\section{Observations}\label{sec:observation}

We present an analysis from both static and dynamic perspectives. Statically, we examine activation maps, while dynamically, we use the unembedding matrix to decode and get the probability distribution over the vocabulary humans understand. 

At the static level, we focus on activation maps.  Figure~\ref{fig:case_activation} shows the top 1\% neurons with the highest activations when the GPT2-XL-1.5B model outputs factual information, resulting in 64 neurons per layer from its 6,400 neurons per layer. Specifically, we test with a labeled factual dataset, sorting the activation values of factual outputs in descending order and retaining the locations of the top 1\% neurons with the highest activations for each data. We calculate the frequency of these neurons and then select those with the highest frequency of appearing in the top 1\% activations. These neurons, which are most frequently highly activated when the LLM outputs factual information, are displayed as factual-related neurons.
Figure~\ref{fig:case_activation}(a) demonstrates that when the LLM provides factual information, the factual-related neurons activate more intensely. This increased activation is likely due to the strengthening of these connections from training on factual data. Conversely, Figure~\ref{fig:case_activation}(b) shows a case of non-factual information with less activation in factual-related neurons. This difference suggests that activation maps can potentially serve as indicators for factual accuracy assessment.

\begin{figure}[t]
    \setlength{\abovecaptionskip}{4pt}
    \setlength{\belowcaptionskip}{0pt}
    \centering
    \includegraphics[width=0.99\linewidth, trim=0 635 60 0,clip]{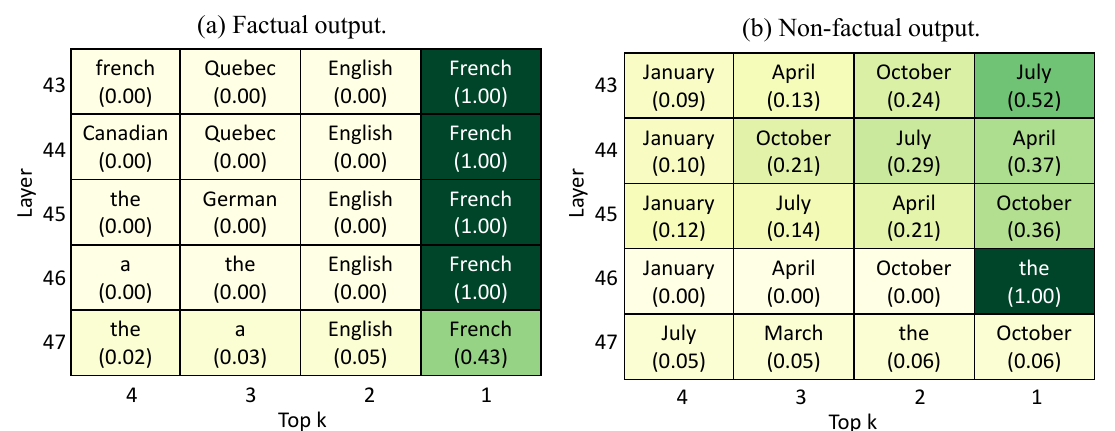}
    \caption{Case study of the top-4 output indices and their probabilities in the last five layers}
    \label{fig:case_topk}
    \vspace{-4mm}
\end{figure}

From the dynamic perspective, we observe the evolution of the final output ranks across layers, depicted in Figure~\ref{fig:case_rank}. 
For instance, when the LLM is queried about the director of ``The Shining'' and initially outputs ``Stanley,'' we observe its rank across layers, where 
rank 1 represents the highest probability. In factual outputs, the final output rank ascends to a higher position after layer 22, showing the LLM's growing confidence to finalize its output in the later stages of the network. In contrast, non-factual outputs show rank fluctuations after layer 35, reflecting the model's uncertainty. 
Further, we observe the top-k output indices and their probabilities. For factual responses, as shown in Figure~\ref{fig:case_topk}(a), the model consistently prioritizes the same word as the highest probability output. On the other hand, while generating non-factual responses, the top probability words are different in the last few layers, as seen in Figure~\ref{fig:case_topk}(b), indicating a lack of confidence. Additionally, the semantic similarity between the top-1 output from one layer to the next is greater for factual than non-factual outputs. These observations reveal that changes in final output rank, top-k output indices, and top-k output probabilities between layers potentially serve as indicators for factual detection.

\section{LLM Factoscope}\label{sec:method}

\begin{figure*}[h]
    \setlength{\abovecaptionskip}{4pt}
    \setlength{\belowcaptionskip}{0pt}
    \centering
    \includegraphics[width=0.8\linewidth, trim=0 585 0 0,clip]{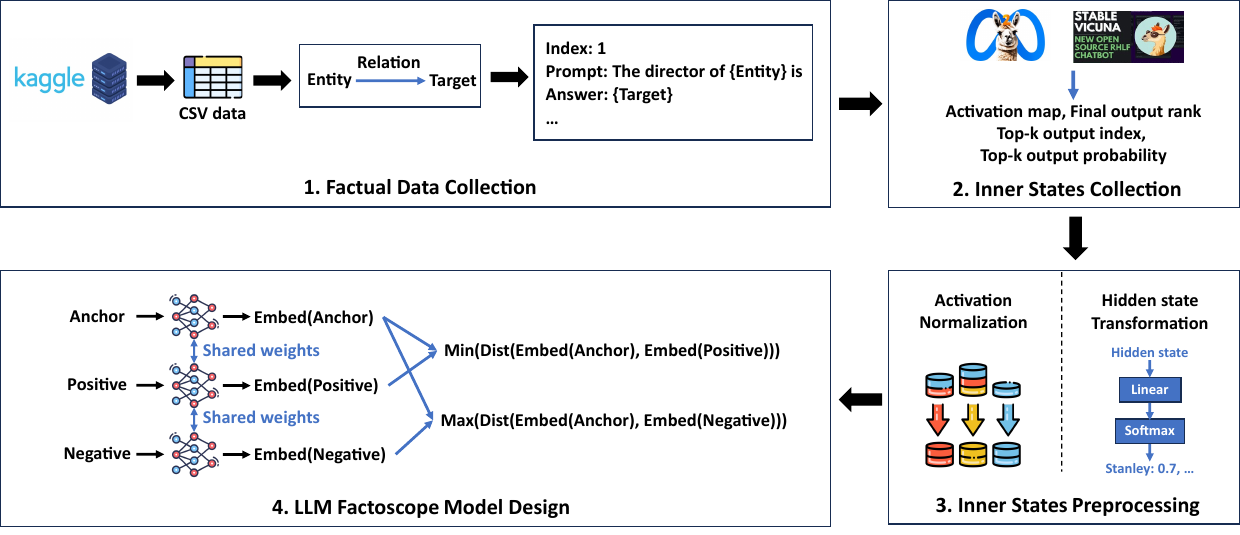}
    \caption{Pipeline of the LLM Factoscope.}
    \label{fig:pipeline}
    \vspace{-4mm}
\end{figure*}

\begin{figure*}[h]
    \setlength{\abovecaptionskip}{4pt}
    \setlength{\belowcaptionskip}{0pt}
    \centering
    \includegraphics[width=0.85\linewidth, trim=20 350 20 0,clip]{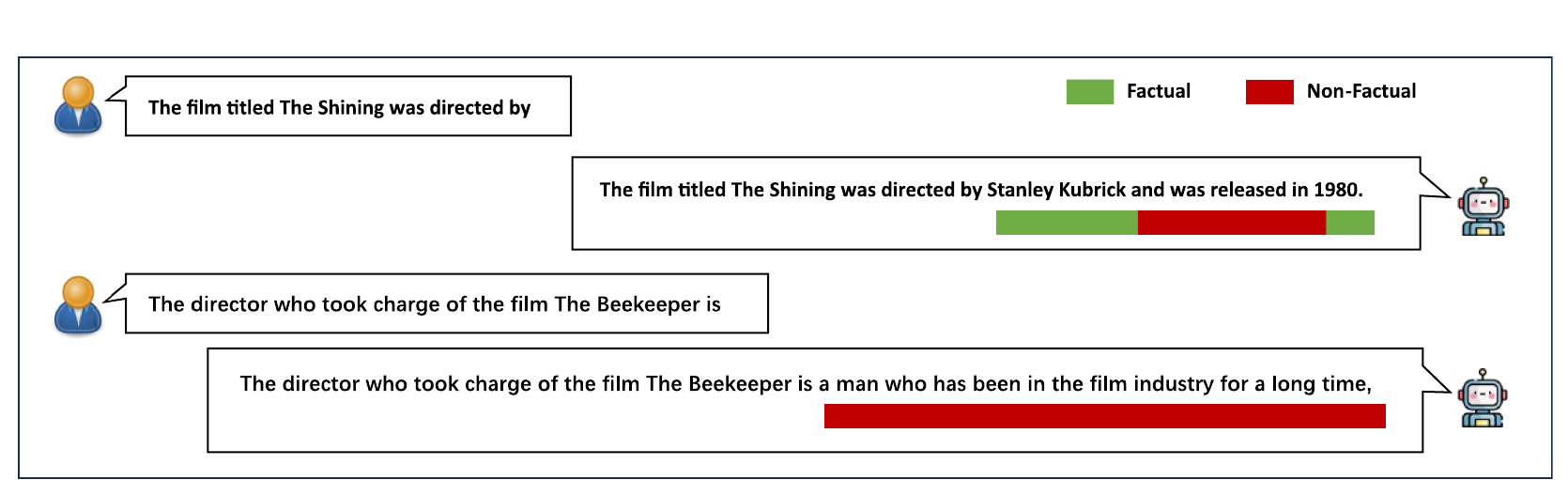}
    \caption{An instance of using LLM Factoscope.}
    \label{fig:instance}
    \vspace{-4mm}
\end{figure*}

In this section, we begin with an overview of our pipeline. Then, we introduce the data collection for factual detection and preprocessing steps for effective model training. Lastly, we present the model architecture of the LLM Factoscope.

\subsection{Overview}

We propose the LLM Factoscope, a pipeline designed to leverage the inner states of LLMs for factual detection, as depicted in Figure~\ref{fig:pipeline}. This pipeline enables easy expansion of the factual detection model's dataset coverage.
Addressing the absence of word-level factual detection methods, we develop our dataset by extracting structured factual data from Kaggle~\cite{kaggle}. 
This dataset is then deployed to probe LLMs to check whether their responses align with factual correctness, serving as labels for our inner states dataset. 
Concurrently, we capture the LLMs' inner states, which include the model's inner representation of knowledge. We use inner states as features for our dataset. 
Following data collection and preprocessing,
we train a Siamese network-based model designed to minimize the embedding distance between similar class data and maximize the distance between pairs of dissimilar class data.

In Figure~\ref{fig:instance}, we show a practical application that enables users to verify the factuality of the LLM’s outputs. Consider a scenario where a user queries an LLM to craft a script for a film introduction video. The user inputs a prompt: ``The film titled The Shining was directed by''. The LLM responds with ``Stanley Kubrick and was released in 1980''. The LLM Factoscope marks the ``Stanley Kubrick'' and the release year ``1980'' as factual. Given another prompt, ``The director who took charge of the film The Beekeeper is'' and the LLM outputs ``a man who has been in the film industry for a long time'' the LLM Factoscope flags each token of the response as non-factual due to the lack of specific factual information about the director's identity or background related to the film mentioned.

\subsection{Factual Data Collection}

\begin{table*}[h]
\setlength{\abovecaptionskip}{4pt}
\setlength{\belowcaptionskip}{0pt}
\scriptsize
\centering
\begin{tabular}{lllr}
\toprule
\textbf{Category} & \textbf{Example} & & \textbf{Size} \\
\midrule
Art~\cite{kaggle_famous_paintings}\cite{kaggle_movies_dataset} & P: The artist of the artwork Still Life with Flowers and a Watch is & A: Abraham Mignon &  67,302 \\
\addlinespace
Sport~\cite{kaggle_olympic_games}\cite{kaggle_top_paying_sports} & P: The athlete Ole Jacob Bangstad represents the country of & A: Norway &  31,718 \\
\addlinespace
Literary~\cite{kaggle_goodreads_books} & P: The book Twilight was written by & A: Stephenie Meyer &  54,301 \\
\addlinespace
Geography~\cite{wikidata_query_service} & P: The city Leipzig is located in the country of  & A: Germany &   1,103 \\
\addlinespace
History~\cite{kaggle_pantheon_project} & P: The birthplace country of the historical figure Albert Einstein is & A: Germany &  56,705 \\
\addlinespace
Science~\cite{kaggle_nobel_laureates} & P: The Nobel laureate Jacobus Henricus van 't Hoff is from & A: Netherlands &   8,971 \\
\addlinespace
Economics~\cite{kaggle_billionaires_statistics} & P: Microsoft was started by & A: Bill Gates &   5,228 \\
\addlinespace
Multi~\cite{ROME} & P: The mother tongue of Danielle Darrieux is & A: French &  21,918 \\ 
\addlinespace
\midrule
Total & & &  247,246 \\
\bottomrule
\end{tabular}
\caption{Overview of the Factual Dataset. Abbreviation: P-Prompt, A-Answer.}
\label{tab:dataset_overview}
\vspace{-2mm}
\end{table*}

We start our dataset collection by searching for factual-related CSV datasets on Kaggle~\cite{kaggle}.
The CSV format's inherent structuring into entities, relations, and targets makes it an ideal candidate for the automated generation of prompts and answers. 
Our dataset includes various categories and
each category includes multiple relations—for instance, in the art category, relationships such as artwork-artist, movie-director, movie-writer, and movie-year are included.
We have manually crafted clear prompts to ensure that LLMs can accurately comprehend the questions, guiding them to directly provide answers, thereby avoiding the repetition of the question. To further enhance the dataset's diversity, we have developed multiple synonymous question templates for each relation. Table~\ref{tab:dataset_overview} provides an overview of the datasets. Users can adopt this collection method with other open datasets to easily expand the coverage of the factual dataset. We include a neuron editing dataset from Meng et al.~\cite{ROME} to enrich the data's scope.

\subsection{Inner States Collection}

After constructing the factual dataset, we feed the prompts into the LLM, capturing the responses and the inner states associated with the last token of each prompt. 
This dataset is designed to verify if the model's next word output aligns with the facts, providing insights into the inner states associated with both factual and non-factual outputs. Due to the autoregressive nature of LLMs, this method enables us to extend factuality predictions to each position in the sequence during testing.
In the following, we detail four key types of inner states: activation maps, final output ranks, top-k output indices and probabilities. 
The first shows the locations and intensity of high activations, representing the static view, while the latter three display the evolution of decision-making and probabilities with a dynamic perspective.

\noindent\textbf{Activation Maps:} For an LLM input sequence $X = \{x_1, x_2, \ldots, x_n\}$, the activation maps $\mathbf{A}_n$ for the last token $x_n$ include the activation values across all layers in the LLM. 
They contain the LLM’s inner representation of the knowledge related to the input. As the LLM traverses through its layers, it retrieves information relevant to the input~\cite{memit}. When the subsequent word aligns with the factual answer, it indicates successful knowledge retrieval at the intermediate layers; otherwise, it suggests inadequate knowledge retrieval. These contrasting scenarios are expected to show distinct activation maps.

\vspace{1mm}
\noindent\textbf{Final Output Ranks:} Let $y$ be the next token output by the LLM in response to the input $X$, and let $H_{l,y}$ represent the hidden state at layer $l$ when outputting $y$. We use $H_{l,y}$  and apply the same vocabulary mapping used in the final hidden layer through a linear and softmax layer, thereby obtaining the probability distribution of vocabulary $V$ at layer $l$, denoted as $P_{l,V}(y|X)$. These distributions are sorted in descending order to get the rank of the final output token $y$ at each layer, symbolized as $R_{l,y}$. The set of ranks $R_{l,y}$ across all layers represents the final output ranks, showing how the position of the final token changes in the probability distribution across different layers. This reflects the model's evolving output preferences and how the likelihood of the final token changes as information is processed through the layers.

\noindent\textbf{Top-k Output Indices and Probabilities:} From the probability distribution $P_{l,V}(y|X)$, we identify the top-k tokens with the highest logits in each layer, represented as $T_{l,k} = argtop_k(P_{l,V}(y|X))$, where $argtop_k$ selects the indices of the top-k highest values in $P_{l,V}(y|X)$. These indices, $T_{l,k}$, represent the model's most likely outputs after processing the information at each layer. With $T_{l,k}$, we then extract the corresponding top-k probabilities $P_{l,k} = \{ P_{l,V}(y|X)[i] \mid i \in T_{l,k} \}$. This data reflects the fluctuating probabilities of these tokens across layers, providing insights into the model's probabilistic reasoning. The relationships among the top-k indices and their probabilities, both within and across layers, shed light on various cognitive aspects of the model’s processing.

Alongside these inner states, we record labels for factual detection. These labels, derived by evaluating whether the model's next word following each prompt aligns with the factual answer, serve as a key indicator of the accuracy in factual detection. A correct alignment is marked as positive, while a misalignment is categorized as negative. 

\subsection{Inner States Preprocessing}

In this section, we introduce the preprocessing of inner states for effective integration into the training process.
We detail the preprocessing methods applied to each category of inner states.

\vspace{1mm}
\noindent\textbf{Normalization of activation map:} We calculate the mean $\mu$ and standard deviation $\sigma$ of the dataset. The activation map $\mathbf{A}_n$ is then normalized using the formula: $\mathbf{A}_{\text{normalized}}=(\mathbf{A}_n-\mu) / \sigma$. This normalization ensures a uniform scale for the activation values, enhancing their comparability and relevance in the model's learning mechanisms.

\vspace{1mm}
\noindent\textbf{Transformation of final output ranks:} We adjust the final output ranks to a 0-1 range, highlighting lower values. Mathematically, the transformation of rank $R_{l,y}$ can be represented as $R_{l,y,\text{transformed}}=1 /[{(1-R_{l,y})+1+10^{-7}}], R_{l,y}\in(1,|V|)$, where $|V|$ is the size of LLM's vocabulary. When the rank $R_{l,y}$ is 1 (indicating the highest rank), the transformed rank $R_{l,y,\text{transformed}}$ becomes its maximum value, close to 1. Adding $10^{-7}$ in the denominator is a small constant to prevent division by zero.

\vspace{1mm}
\noindent\textbf{Distance calculation for top-k output indices:} In processing the top-k output indices, we measure the semantic similarity across adjacent layers by calculating the cosine similarity between the embeddings of tokens, providing insights into how the model's decision-making and semantic continuity evolves across layers. 

\vspace{1mm}
It is important to note that while the above inner states require preprocessing to standardize their scales or enhance their interpretability, the top-k output probabilities do not need such preprocessing. This is because the top-k output probabilities are inherently on a consistent scale, being probabilities that naturally range from 0 to 1. 

\subsection{LLM Factoscope Model Design}

After preparing the dataset of inner states, we develop the LLM Factoscope model, inspired by the principles of few-shot learning and Siamese networks. It is designed to effectively learn robust representations from limited data. This approach distinguishes between factual and non-factual content and demonstrates impressive generalization capabilities.
Our model comprises four distinct sub-models, each processing one of the inner states.

For the activation maps, top-k output indices, and top-k output probabilities, we utilize Convolutional Neural Networks (CNNs) with the ResNet18 architecture~\cite{resnet}. The choice of ResNet18, with its convolutional and residual connections, is particularly advantageous for efficiently capturing the relationships between and within different layers of the LLM. These CNNs transform the inner states into embeddings $\mathbf{E}_{\text{activation}}$, $\mathbf{E}_{\text{top-k index}}$, and $\mathbf{E}_{\text{top-k prob}}$. Each embedding captures unique aspects of the LLM's processing dynamics. As for the final output ranks, a sequential data type, we use a Gated Recurrent Unit (GRU) network~\cite{gru}, reflecting the temporal evolution of the model's output preferences across layers. This network yields an embedding \( \mathbf{E}_{\text{rank}} \). The embeddings from these four sub-models are then integrated through a linear layer to form a comprehensive mixed representation, $\mathbf{E}_{\text{mixed}}$, which captures an integrated expression of the LLM's factual understanding, representing spatial and temporal insights.


During training, our model uses the triplet margin loss~\cite{triplet_margin_loss}, a metric integral to embedding learning in few-shot learning scenarios. Florian\etal~\cite{facenet} demonstrated that triplet margin loss can significantly improve generalization across varied tasks. When our model encounters OOD data and its performance drops, it can adapt swiftly and cost-effectively by adding minimal additional support data. In contrast, regular classifiers require extensive data and retraining, leading to higher costs and reduced flexibility. This loss function minimizes the distance between instances of the same class while maximizing the distance between instances of different classes. Each training instance serves as an anchor. If the anchor is a factual instance, the positive example is another factual instance, and the negative example is a non-factual instance. Conversely, if the anchor is a non-factual instance, the positive example is another non-factual instance, and the negative example is a factual instance.

For a given training instance $x$, we feed it to the LLM Factoscope model and get an embedding for its mixed representation, $\mathbf{E}_{\text{anchor}}$. Then, we select a positive example $\mathbf{x}_{\text{pos}}$ from the same category as the anchor and a negative example $\mathbf{x}_{\text{neg}}$ from a different category. Subsequently, we obtain their respective mixed expressions $\mathbf{E}_{\text{pos}}$ and $\mathbf{E}_{\text{neg}}$.
The triplet margin loss aims to ensure that the distance between the anchor and the positive instance, $\text{Dist}(\mathbf{E}_{\text{anchor}}, \mathbf{E}_{\text{pos}})$, is smaller than the distance between the anchor and the negative instance, $\text{Dist}(\mathbf{E}_{\text{anchor}}, \mathbf{E}_{\text{neg}})$, by at least a margin $\alpha$. This loss function is formally defined as:
$L=\max (\text{Dist}(\mathbf{E}_{\text{anchor}}, \mathbf{E}_{\text{pos}}) - \text{Dist}(\mathbf{E}_{\text{anchor}}, \mathbf{E}_{\text{neg}}) + \alpha, 0),$
where $\text{Dist}(\cdot,\cdot)$ is the chosen distance metric, typically Euclidean distance. By fine-tuning $\alpha$, we can enhance the model's discriminative capability, ensuring that the distance between the anchor and the positive instance is less than that between the anchor and the negative instance by at least the margin $\alpha$. The training process minimizes the loss, refining the model's ability to accurately differentiate between factual and non-factual content.

In the testing phase, we establish a support set consisting of data samples and their corresponding targets, denoted as $\{ S_{1}, \ldots, S_{n} \}$ and $\{ T_{\text{sup}_1}, \ldots, T_{\text{sup}_n} \}$, respectively. These samples have not been used in the training process of the LLM Factoscope model. They provide a reference for comparing and classifying new, unseen test data. Each sample in the support set is processed through the LLM Factoscope model to generate mixed representations, represented by $\{\mathbf{E}_{\text{sup}_1}, \ldots, \mathbf{E}_{\text{sup}_n}\}$. The mixed representations are outputs of the LLM Factoscope model. The test data's mixed representation, $\mathbf{E}_{\text{test}}$, is then compared against these support set representations.
The classification of the test data is determined by identifying the closest support set embedding to $\mathbf{E}_{\text{test}}$. The target of the test data is the target of this nearest support set data: 
$T_{\text{test}} = T_{\text{sup}_{i^*}} \quad \text{where} \quad i^* = \argmin_{i} \text{Dist}(\mathbf{E}_{\text{test}}, \mathbf{E}_{\text{sup}_i}).$
Here, the index $i^*$ identifies the support set data that is closest to the test data, and $T_{\text{sup}_{i^*}}$ is the target associated with this closest support set data. 

\section{Evaluation}


\subsection{Experimental Setup}\label{subsec:setup}

\begin{table*}[ht]
\setlength{\abovecaptionskip}{4pt}
\setlength{\belowcaptionskip}{0pt}
\scriptsize
\centering
\caption{Effectiveness results of LLM Factoscope across different LLM architectures.}
\label{tab:effectiveness}
\begin{tabular}{lcccccc}
\toprule
Method & GPT2-XL-1.5B & Llama2-7B & Vicuna-7B & Stablelm-7B & Llama2-13B & Vicuna-13B \\
\midrule
Ours & \textbf{0.961} & \textbf{0.967} & \textbf{0.982} & \textbf{0.983} & \textbf{0.983} & \textbf{0.974} \\
Baseline & 0.880 & 0.888 & 0.831 & 0.817 & 0.882 & 0.785 \\
SAPLMA~\cite{internal} & 0.861 & 0.843 & 0.863 & 0.801 & 0.854 & 0.839 \\
Calibration-Prob~\cite{calibration} & 0.868 & 0.715 & 0.725 & 0.864 & 0.657 & 0.709 \\
Calibration-LLM Label~\cite{calibration} & 0.105 & 0.344 & 0.391 & 0.347 & 0.405 & 0.373 \\
SelfCheckGPT~\cite{selfcheckgpt} & 0.646 & 0.568 & 0.623 & 0.645 & 0.551 & 0.634 \\
\bottomrule
\end{tabular}
\vspace{-2mm}
\end{table*}

\begin{table*}[ht]
\setlength{\abovecaptionskip}{4pt}
\setlength{\belowcaptionskip}{0pt}
\scriptsize
\centering
\caption{Generalization Performance of different LLM architectures. Abbreviations: BL - Baseline, BA - Book-Author, CC - Celebrity-Country, AC - Athlete-Country.}
\label{tab:generalization}
\begin{tabular}{lccccccccccccc}
\toprule
\multirow{3}{*}{Data} & \multicolumn{3}{c}{GPT2-XL-1.5B} & \multicolumn{3}{c}{Llama2-7B} & \multicolumn{3}{c}{Vicuna-7B} \\
\cmidrule(r){2-4} \cmidrule(lr){5-7} \cmidrule(l){8-10} 
& Ours & BL & SAPLMA & Ours & BL & SAPLMA & Ours & BL & SAPLMA \\
\midrule
BA & 0.712 & 0.800 & \textbf{0.836} & \textbf{0.977} & 0.701 & 0.770 & \textbf{0.971} & 0.757 & 0.760 \\
CC & 0.871 & \textbf{0.879} & 0.747 & \textbf{0.972} & 0.946 & 0.759 & 0.790 & 0.516 & \textbf{0.855} \\
AC & \textbf{0.979} & 0.780 & 0.720 & \textbf{0.703} & 0.693 & 0.635 & \textbf{0.770} & 0.716 & 0.734 \\
\midrule
Average & \textbf{0.854} & 0.818 & 0.768 & \textbf{0.884} & 0.780 & 0.721 & \textbf{0.844} & 0.663 & 0.750 \\
\midrule
\midrule
\multirow{3}{*}{Data} & \multicolumn{3}{c}{Stablelm-7B} & \multicolumn{3}{c}{Llama2-13B} & \multicolumn{3}{c}{Vicuna-13B} \\
\cmidrule(lr){2-4} \cmidrule(l){5-7} \cmidrule(l){8-10}
& Ours & BL & SAPLMA & Ours & BL & SAPLMA & Ours & BL & SAPLMA \\
\midrule
BA & 0.690 & 0.333 & \textbf{0.704} & \textbf{0.904} & 0.854 & 0.830 & \textbf{0.895} & 0.814 & 0.776 \\
CC & 0.635 & 0.610 & \textbf{0.705} & \textbf{0.938} & 0.849 & 0.608 & \textbf{0.913} & 0.698 & 0.702 \\
AC & 0.694 & 0.756 & \textbf{0.728} & 0.778 & 0.686 & \textbf{0.822} & \textbf{0.807} & 0.703 & 0.817 \\
\midrule
Average & 0.673 & 0.566 & \textbf{0.712} & \textbf{0.873} & 0.763 & 0.753 & \textbf{0.872} & 0.738 & 0.765 \\
\bottomrule
\end{tabular}
\vspace{-4mm}
\end{table*}

\textbf{Task.} Our task is to train an LLM Factoscope model for detecting the factuality of LLMs' outputs. The inputs of the LLM Factoscope are the inner states of the last token of the LLM's inputs. The output is a binary label, indicating whether the LLM's output token is factual or not. We input the prompt of our dataset (\eg ``The birthplace country of the historical figure Albert Einstein is'') and then check if the first output word of the LLM aligns with the fact (\eg ``Germany''). If it does, the LLM's output is classified as factual. If the first output word consists of multiple tokens, we will concatenate them to form a complete word.

\vspace{1mm}
\noindent
\textbf{Dataset.} We employ various factual datasets encompassing art, sports, literature, geography, history, science, and economics, comprising 247,246 data points.
Then, we record the inner states of the LLM as it provides factual and non-factual outputs, including activation maps, final output ranks, top-k output indices, and top-k output probabilities.
The label assigned to each data point indicates whether the corresponding model output is factual or non-factual. 
To ensure dataset balance, we randomly select an equal number of factual and non-factual data points for each factual relation. The dataset is preprocessed to ensure it is standardized and optimized for model learning.

\vspace{1mm}
\noindent
\textbf{Models.} Our experiments are conducted on several popular LLMs, each with distinctive architectures and characteristics. These models include GPT2-XL~\cite{gpt2xl}, Llama-2-7B~\cite{llama}, Llama-2-13B, Vicuna-7B-v1.5~\cite{vicuna}, Vicuna-13B-v1.5, and StableLM-7B~\cite{alpaca}. 
These models allow us to comprehensively evaluate the effectiveness of our factual detection method across various LLM architectures and configurations.

\vspace{1mm}
\noindent
\textbf{Baselines. }
We compare LLM Factoscope against five existing methods: two white-box methods and three black-box methods. The first white-box method is a simple baseline that uses the activations from a single layer of the LLM, while the second, SAPLMA~\cite{internal}, uses hidden states from a single layer to detect the factuality of input sentences. We trained SAPLMA on our dataset to adapt it for detecting LLM output factuality.
The three black-box methods rely on the LLM's final output and probabilities. Calibration-Prob~\cite{calibration} uses output probabilities to determine the factuality of generated content, Calibration-LLM Label~\cite{calibration} relies on the LLM to label whether the output is factual, and SelfCheckGPT~\cite{selfcheckgpt} uses repetitive sampling to assess output uncertainty.


\subsection{Effectiveness}

To evaluate the effectiveness of the LLM Factoscope, we compared it against five existing methods. 
Detailed settings are shown in Appendix~\ref{sec:training_details}.
As shown in Table~\ref{tab:effectiveness}, our LLM Factoscope consistently maintains high accuracy levels, ranging between 96.1\% and 98.3\%, across different LLM architectures. In contrast, the accuracy of the baseline and SAPLMA fluctuates between 78.5\% and 88.8\%. This variation suggests that as LLMs increase in parameter size, the regions responsible for different types of factual knowledge might differ, or multiple layers could be involved in representing a single type of factual knowledge. Consequently, the baseline, which relies solely on activation values and hidden states from a single layer, demonstrated unstable performance.
The three black-box methods demonstrated lower effectiveness compared to the first two white-box methods due to their limited accessible information. Among the black-box methods, Calibration-Prob performed the best.

Based on the analysis, we believe that the superior performance of the LLM Factoscope can be attributed to its consideration of various inner state changes across layers. By integrating this multi-dimensional analysis of inner states within LLMs, LLM Factoscope effectively discerns factual from non-factual outputs, offering a more robust and reliable approach to factual detection. This method's success not only highlights the significance of activation values in understanding LLM outputs but also paves the way for future explorations into the intricate workings of LLMs, particularly in the realm of natural language processing applications. We also provide an interpretation analysis on the LLM Factoscope in Appendix~\ref{sec:interpretation}.

\begin{figure*}[htbp]
\setlength{\abovecaptionskip}{4pt}
\setlength{\belowcaptionskip}{0pt}
\centering
\begin{minipage}[t]{0.35\textwidth}
\centering
\includegraphics[width=0.57\textwidth, trim=0 12 0 25,clip]{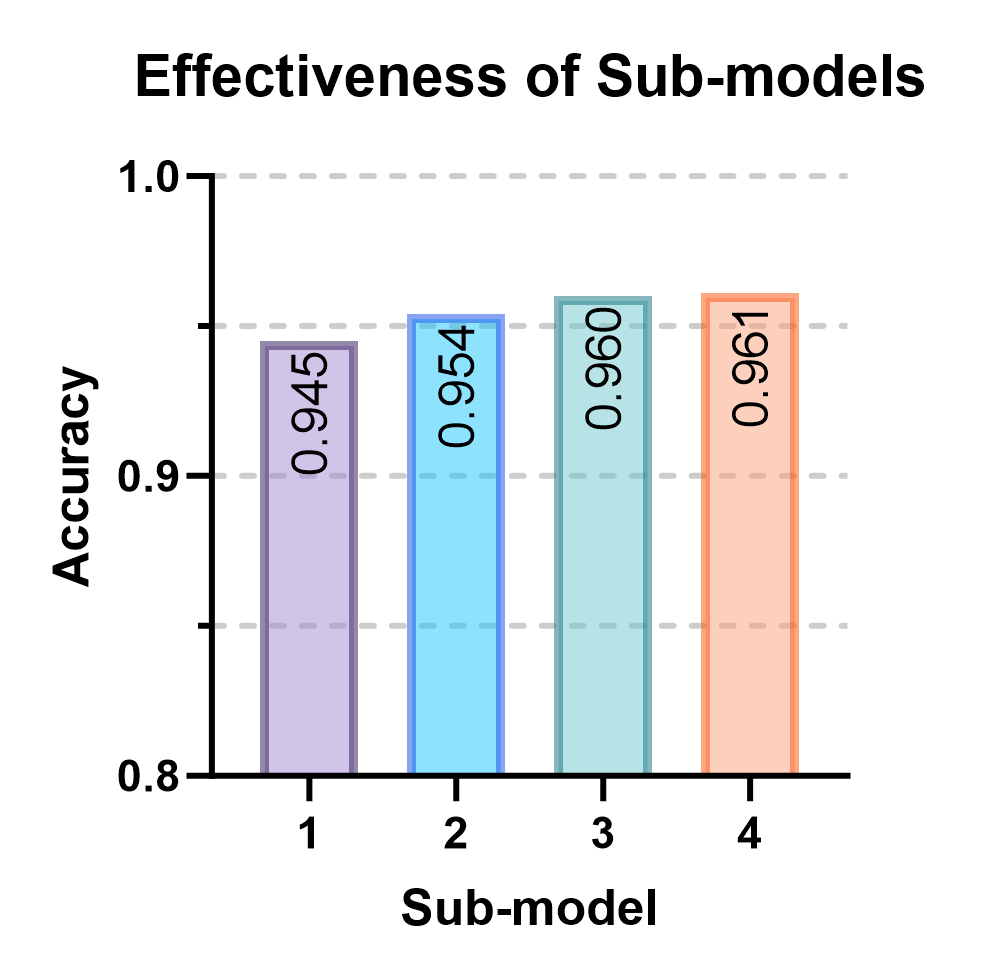}
\caption{Effectiveness of Sub-models}
\label{fig:effective_submodel}
\end{minipage}
\begin{minipage}[t]{0.45\textwidth}
\centering
\includegraphics[width=0.99\textwidth, trim=0 10 0 25,clip]{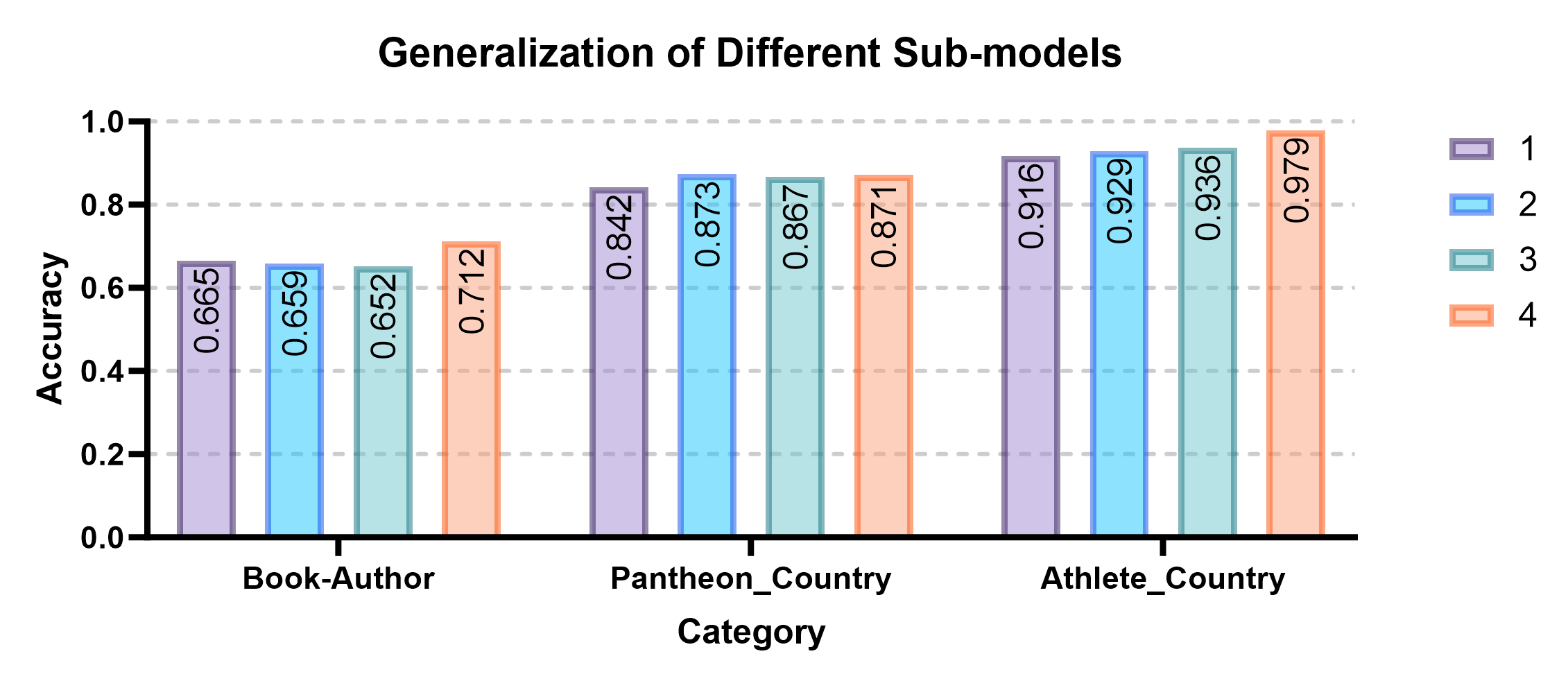}
\caption{Generalization of Sub-models}
\label{fig:generalization_submodel}
\end{minipage}
\vspace{-2mm}
\end{figure*}

\begin{figure*}[htbp]
\setlength{\abovecaptionskip}{4pt}
\setlength{\belowcaptionskip}{0pt}
\centering
\begin{minipage}[t]{0.25\textwidth}
\centering
\includegraphics[width=0.85\textwidth, trim=0 10 0 27,clip]{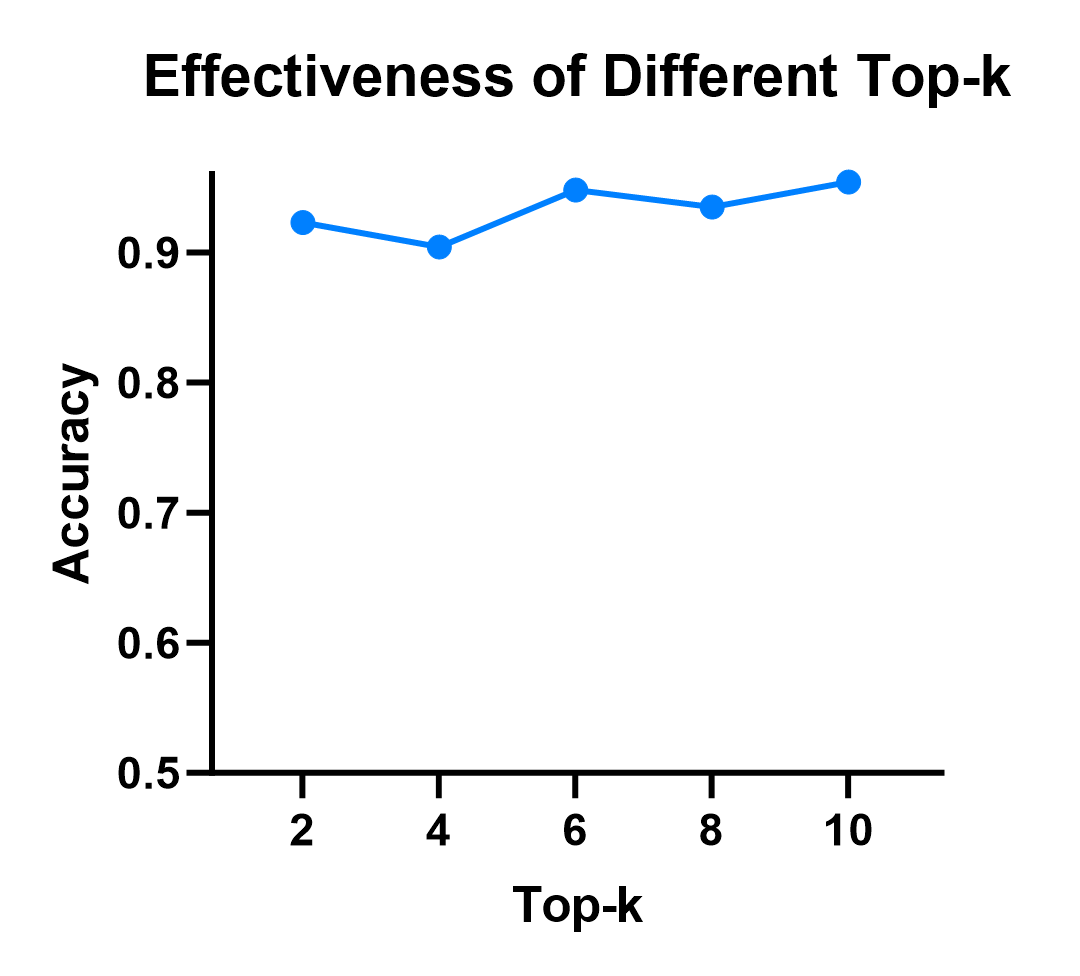}
\caption{Effects of Top-k}
\label{fig:topk}
\end{minipage}
\begin{minipage}[t]{0.33\textwidth}
\centering
\includegraphics[width=0.85\textwidth, trim=0 10 0 27,clip]{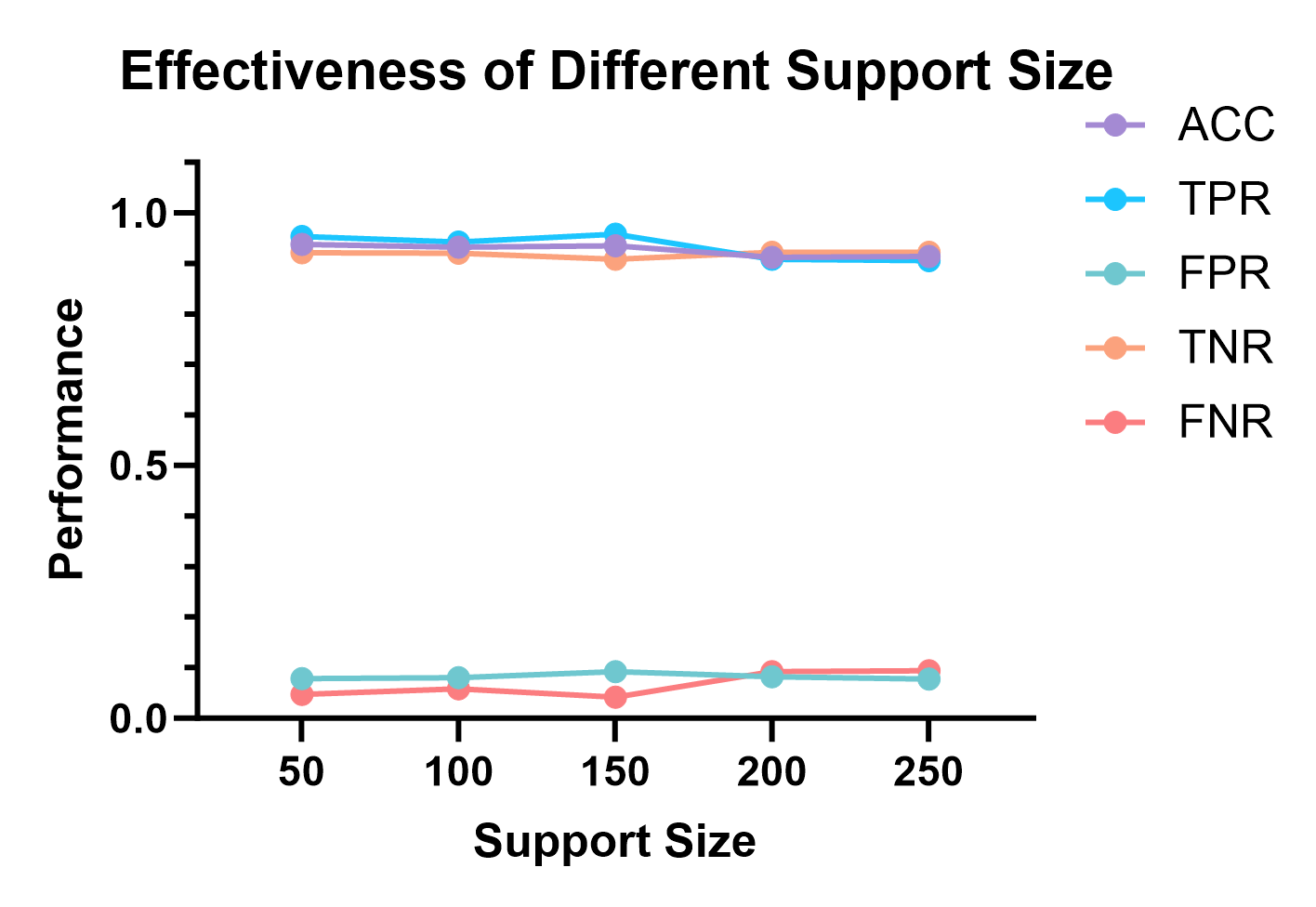}
\caption{Effects of support set size}
\label{fig:supportsize}
\end{minipage}
\begin{minipage}[t]{0.29\textwidth}
\centering
\includegraphics[width=0.7\textwidth, trim=0 10 0 25,clip]{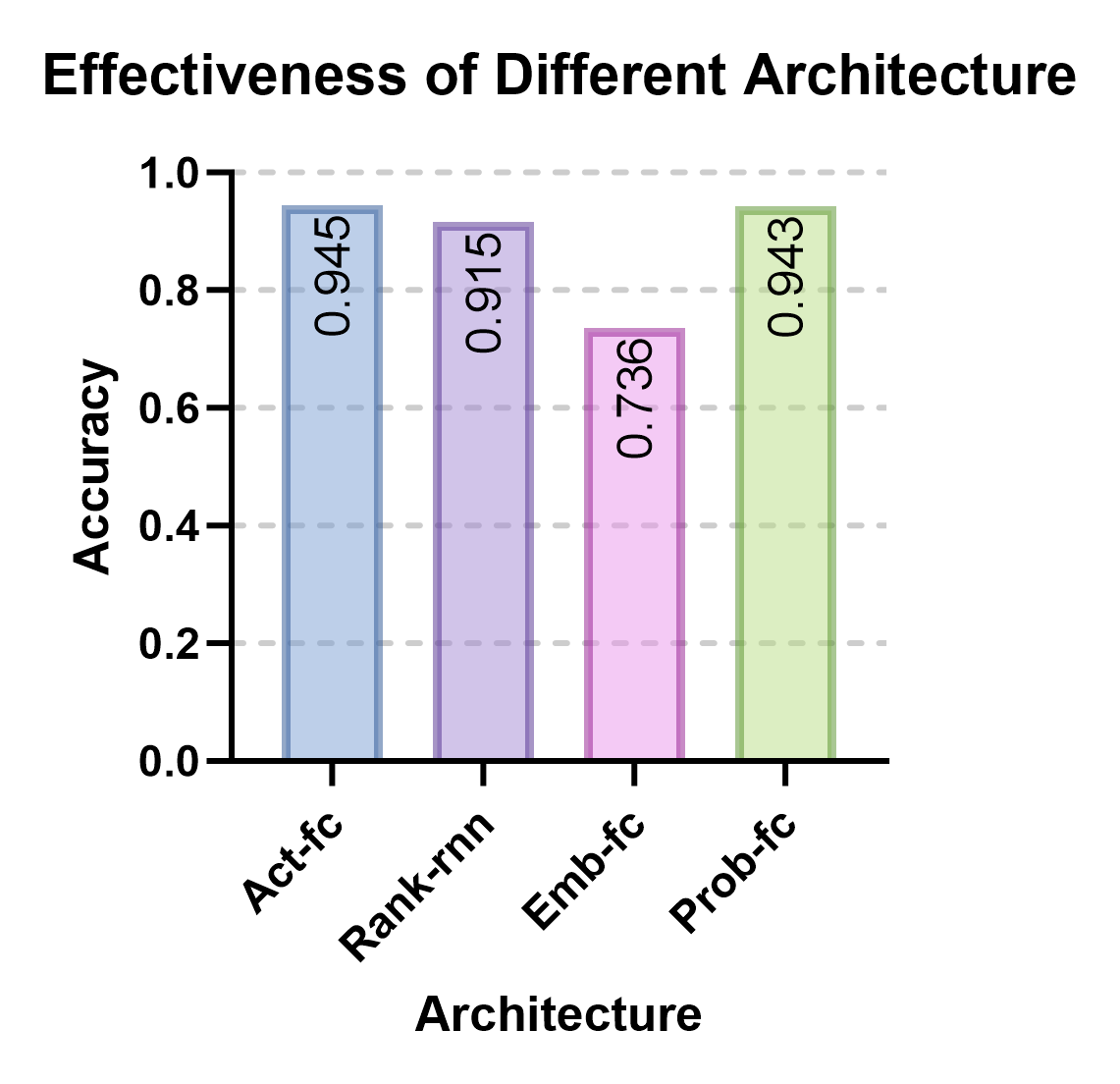}
\caption{Effects of architecture}
\label{fig:arch}
\end{minipage}
\vspace{-4mm}
\end{figure*}

\subsection{Generalization}\label{subsec:generaliztion}

It is well-established in neural network research that the effectiveness of a model largely depends on the similarity between training and testing distributions~\cite{oodsurvey}. Thus, our model's performance may vary across different distributions. We adopt a leave-one-out approach for our generalization assessment. Specifically, we remove one relation dataset, train the model on the remaining datasets, and then test it on the omitted dataset. We selected three relations for assessing generalization, including Book-Author, Celebrity-Country, and Athlete-Country, as these relations form sizable datasets across all LLMs tested. We also compare our method with baseline and SAPLMA that train models for factual detection.

Different LLMs exhibit varying generalization capabilities across different relations, as shown in Table~\ref{tab:generalization}. In the ``Book-Author'' relation, our method achieves a notable 97.7\% accuracy with Llama2-7B but 69.0\% with Stablelm-7B. This variation is likely due to each LLM's unique handling of different types of factual knowledge. Our method outperforms the baselines in most cases, with its average performance outperforming the baseline except for Stablelm-7B.
We hypothesize that this could be attributed to its less effective learning of factual versus non-factual content it was trained on.
The Stablelm-7B model exhibits unstable variations in the semantic continuity of its top-k outputs when predicting OOD data, which may be attributed to the fewer layers it has. This could compromise its generalization ability to maintain semantic continuity across top-k outputs and significantly impact its performance during OOD testing.

The activation maps are sensitive to the data type being processed; for instance, neurons activated in response to art-related prompts may differ from those responding to science-related prompts. In contrast, final output ranks, top-k output indices, and top-k output probabilities are agnostic to the data type. Leveraging these features to observe the LLM's decision-making evolution significantly enhances the generalization capabilities of the LLM Factoscope across diverse domains.
While our LLM Factoscope shows generalization, we advise matching testing and training data distributions for optimal use. For example, an LLM used as a historical assistant should train the LLM Factoscope with historical data for best performance.

\subsection{Ablation Study}\label{subsec:ablation}




\noindent
\textbf{Contribution of each sub-model.} We evaluate the contribution of each sub-model by incrementally adding them to the factual detection model on GPT2-XL-1.5B. As depicted in Figure~\ref{fig:effective_submodel}, we notice a slight but consistent improvement in accuracy with adding more sub-models. This indicates that each sub-model brings a unique dimension to the model's capabilities, enhancing its overall performance. We employ the ``leave one out'' training approach as in Section~\ref{subsec:generaliztion} to assess the contribution of sub-models to generalization. The results in Figure~\ref{fig:generalization_submodel} demonstrate enhanced generalization as more sub-models are integrated. 
The figures illustrate a sequential integration, starting with ``1'' that uses activation maps, and culminating in ``4'' that combines activation maps, final output ranks, top-k output indices, and top-k output probabilities.
This improvement is particularly evident in the final model, which shows an increase in accuracy across various datasets compared to the model with only one sub-model. 

\vspace{1mm}
\noindent
\textbf{Effects of different top-k.} The top-k affects the top-k output indices and top-k output probabilities. The previous experiments set the top-k to top-10 unless otherwise stated. Now, we will evaluate the effect of choosing different values for top-k on the performance of the factual detection model. We set the k to 2, 4, 6, 8, 10 on GPT2-XL-1.5B. The results of the experiment are shown in Figure~\ref{fig:topk}. The lowest performance is 90.4\% when k is 4, and the highest performance is 95.4\% when k is 10. The difference between the two is 5\%.


\vspace{1mm}
\noindent
\textbf{Effects of different support set size.} We also try different support set sizes from 50 to 250, observing their impact on the performance of the factual detection model. 
This evaluation was conducted on the Llama2-7B. 
The results, as presented in Figure~\ref{fig:supportsize}, demonstrate that the change in support set size does not significantly impact the model's performance across most metrics. 



\vspace{1mm}
\noindent
\textbf{Effects of different sub-models' architectures.} We use different sub-model architectures and assess the performance of the factual detection model. We use fully connected layers to replace the ResNet18 and RNN to replace the GRU network. As shown in Figure~\ref{fig:arch}, 
when we replace parts of the architecture with fully connected layers (act-fc, prob-fc) and RNNs (rank-rnn), we notice a slight decline in performance. 
In contrast, the emb-fc architecture, where we replace the ResNet18 with fully connected layers, results in a significant performance drop
with accuracy falling to 73.6\%. Such a drastic drop highlights the pivotal role of ResNet18 in effectively capturing the LLM’s top-k output indices.
The results show that while the model demonstrates resilience to certain architectural changes, some alterations can substantially impact its performance.



\section{Conclusion}

We discover effective inner states and provide observation for LLM factual detection. Then, we develop a factual detection pipeline, the LLM Factoscope, which leverages these inner states to detect the factuality of LLM outputs. The LLM Factoscope consistently demonstrated high factual detection accuracy, surpassing 96\% in our custom-collected datasets. Our research not only provides a novel method for LLMs' factual detection but also opens new avenues for future explorations into the LLMs' inner states. By paving the way for enhanced model understanding and reliability, the LLM Factoscope sets a foundation for more transparent, accountable, and trustworthy use of LLMs in critical applications.



\vspace{2mm}
\noindent\textbf{Limitations.} 
Our research assumes LLMs' training corpora are predominantly factual, despite the potential inclusion of some non-factual content. Our pipeline remains effective as long as factual outweighs non-factual data. 
When creating our dataset, we generate various prompts and responses to simulate factual inquiries. Despite our efforts, the diversity and complexity of real-world knowledge may not be fully represented in our custom dataset. However, owners of LLMs providing factual detection services likely have access to more comprehensive and meticulously vetted datasets. Such resources could potentially enhance the performance and accuracy of LLM Factoscope.
The LLM Factoscope shows generalizability in our evaluations, effectively detecting factuality across various contexts. However, it may encounter challenges when dealing with data types that significantly deviate from those in our training dataset. Following our pipeline to fine-tune the model with new, more representative datasets could further refine its accuracy and broaden its applicability.


\vspace{2mm}
\noindent\textbf{Ethical Consideration.} Our research utilizes publicly available datasets from Kaggle, which do not contain sensitive content, thereby mitigating direct concerns regarding data privacy and protection. We have adhered strictly to Kaggle's terms of use and conditions, ensuring the lawful utilization of these datasets. Although our research aims to detect the factuality of LLMs' outputs, the development and application of this technology could have broader social impacts. Positively, improving information accuracy can increase public trust in digital content and has beneficial implications for education, scientific research, and journalism by verifying the factuality of LLM outputs. However, we are also cautious of potential misuse, such as employing this technology for censorship, controlling information flow, exacerbating subjective interpretations and manipulation of ``facts''.

\section*{Acknowledgements}
We thank all the anonymous reviewers for their constructive feedback.
The IIE authors are supported in part by NSFC (Grant No.92270204, 62302497 and 62302498), Youth Innovation Promotion Association CAS and China Science and Technology Cloud.

\bibliography{ref}


\appendix

\section{Training Details}
\label{sec:training_details}

\noindent
\textbf{Dataset. }
The training set to test set ratio is 0.8:0.2. Due to varying numbers of factual and non-factual outputs for each LLM, we employ datasets of different scales to train the Factoscope for different LLMs, as shown in Table~\ref{tab:llm_data}.
We use a Siamese-based model architecture in the LLM Factoscope. The LLM Factoscope model comprises several sub-models, each tailored to handle a specific type of inner state. This includes a ResNet18 model for processing activation values, a GRU network for final output rankings, and two additional ResNet18 models for handling top-k output indices and top-k output probabilities. 

\vspace{2mm}
\noindent
\textbf{Parameters. }
We set top-k to top-10. The output of each sub-model is an embedding of dimension 24. These embeddings from each sub-model are concatenated, resulting in a combined embedding of dimension 96. (24 dimensions from each of the four sub-models). This combined embedding is then fed into a fully connected layer, which reduces the dimensionality to 64, ensuring a compact yet informative representation. The final embedding undergoes ReLU activation and L2 normalization, providing a normalized feature vector for each input. During testing, the size of the support set is set to 100. During training, the triplet margin loss function with a margin $\alpha$ of 1.0. This configuration enhances the model’s distinction between factual and non-factual content. 

We use activation values from specific layers to establish a comparative baseline.
For GPT2-XL-1.5B, the model is based on the activation values from the 31st layer. In the case of Llama2-7B and Vicuna-7B, the 23rd layer's activation values are used. For Stablelm-7B, the baseline model relied on the 12th layer, while Llama2-13B and Vicuna-13B utilize the activation values from their 32nd layers.
To ensure reproducibility and transparency in our research, we have documented the key parameters relevant to our experiments in Table~\ref{tab:model_parameters}. This detailed listing of hyperparameters aims to provide comprehensive insights into the configuration settings used in our model, thereby facilitating the replication of our study and validation of our findings.

\vspace{2mm}
\noindent
\textbf{Environment. }
We conduct experiments on a server with 32 Intel Xeon Silver 4314 CPUs at 2.40 GHz, 386 GB of RAM, and an NVIDIA A100 Tensor Core GPU. Training LLM Factoscope on different LLM architectures takes under 24 hours. Results presented are from a single run, but multiple runs yield similar outcomes.

\begin{table}[t]
\scriptsize
\centering
\caption{Training and Testing Data for Various LLM Architectures}
\label{tab:llm_data}
\begin{tabular}{lcc}
\toprule
Model & Train & Test \\
\midrule
GPT2-XL-1.5B & 21,254 & 9,009 \\
Llama2-7B & 57,614 & 24,593 \\
Llama2-13B & 49,282 & 21,022 \\
Vicuna-7B & 51,949 & 22,164 \\
Vicuna-13B & 50,936 & 21,730 \\
Stablelm-7B & 3,445 & 1,377 \\
\bottomrule
\end{tabular}
\end{table}

\begin{table}[h]
\scriptsize
\centering
\begin{tabular}{|c|c|}
\hline
\textbf{Parameter} & \textbf{Value} \\
\hline
k & 10 \\
Sub-model Output Embedding Dimension & 24 \\
Combined Embedding Dimension & 96 \\
Fully Connected Layer Output Dimension & 64 \\
Activation Function & ReLU \\
Normalization & L2 normalization \\
Support Set Size & 100 \\
Epochs & 30 \\
Batch Size & 64 \\
GRU Model Hidden Dimensions & 1st - 128, 2nd - 64 \\
Dataset Split & 80\% training, 20\% testing \\
Activation Dimension of GPT2-XL & 6400 \\
Activation Dimension of Llama-2-7B & 11008 \\
Activation Dimension of Vicuna-7B-v1.5 & 11008 \\
Activation Dimension of Llama-2-13b & 13824 \\
Activation Dimension of StableLM-7B & 24576 \\
Optimizer & Adam \\
Learning Rate & 0.001 \\
Betas & 0.9, 0.999 \\
Epsilon & 1e-08 \\
Weight Decay & 0 \\
Loss Function & Triplet Margin Loss \\
Margin & 1.0 \\
p & 2.0\\
Epsilon & 1e-06 \\
\hline
\end{tabular}
\caption{Model Parameters for LLM Factoscope}
\label{tab:model_parameters}
\vspace{-2mm}
\end{table}

\section{Interpretability}\label{sec:interpretation}

\begin{figure}[t]
    \setlength{\abovecaptionskip}{4pt}
    \setlength{\belowcaptionskip}{0pt}
    \centering
    \includegraphics[width=0.88\linewidth, trim=0 350 170 0,clip]{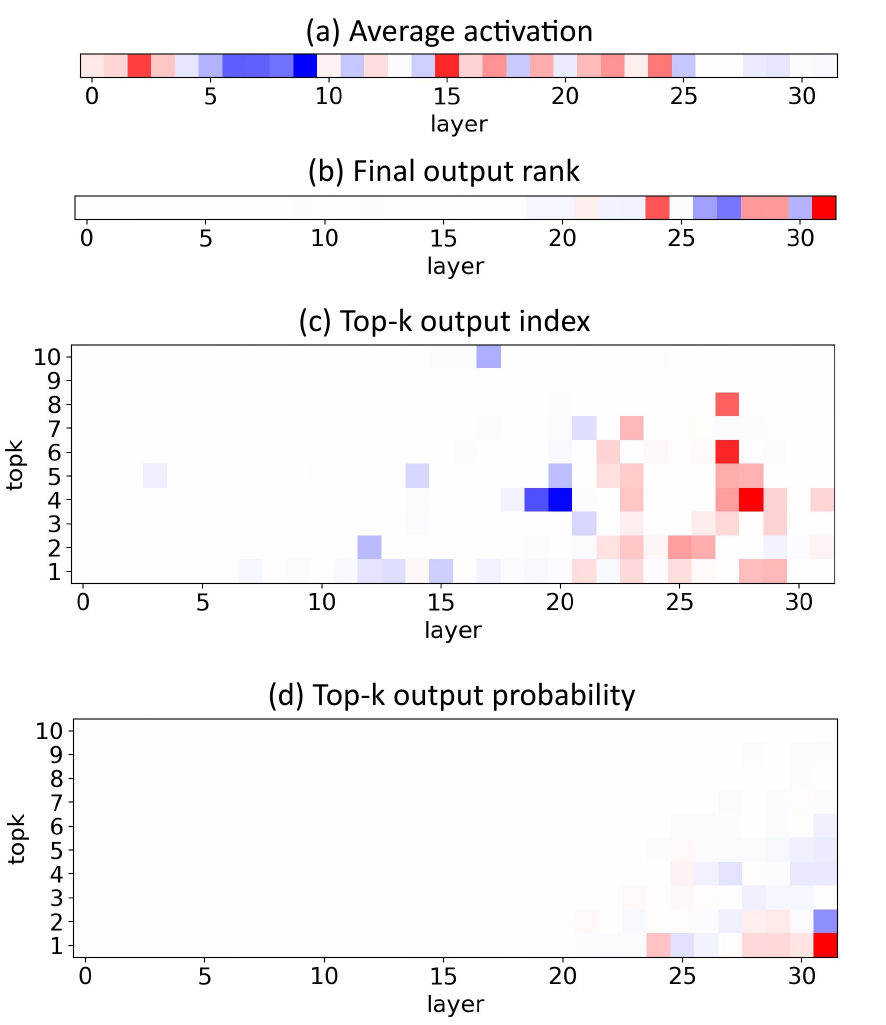}
    \caption{Visualization of different inner states' contribution.}
    \label{fig:interpretation}
    \vspace{-4mm}
\end{figure}

We delve into the interpretability of the LLM Factoscope, aiming to analyze the contribution of these features in discerning the factualness of LLM outputs.
Specifically, we use the Integrated Gradient~\cite{ig} to analyze the contribution of activation maps, final output ranks, top-k output indices, and top-k output probabilities. Integrated Gradient is particularly chosen for its higher faithfulness in interpretability assessments~\cite{trendtest}.
Our analysis reveals that the most influential features are mainly in the middle to later layers of the LLMs, consistently observed across all four data types. To provide a clearer visualization of this pattern, we present a typical example in Figure~\ref{fig:interpretation}. In the figure, red indicates a positive contribution, while blue signifies a negative contribution, with deeper colors representing higher importance. Due to the high dimensionality of activation maps, we display the average importance of features at each layer. 
The majority of positive contributions emerge after the 15th layer. This finding aligns with our observations that the model initially filters semantically coherent candidate outputs in the earlier layers, and then progressively focuses on candidates relevant to the given prompt task in deeper layers.

\end{document}